\pdfoutput=1

\documentclass[11pt]{article}

\usepackage{acl}

\usepackage{times}
\usepackage{latexsym}

\usepackage[T1]{fontenc}

\usepackage[utf8]{inputenc}

\usepackage{microtype}

\usepackage{graphicx}

\usepackage{csquotes}

\usepackage{amsthm}
\theoremstyle{definition}
\newtheorem{exmp}{Example}

\usepackage{enumitem}
\usepackage{multirow}
\usepackage{array}
\usepackage{booktabs}
\usepackage{cellspace}
\usepackage{tabularx}
\usepackage{makecell}

\usepackage{amsmath}

\usepackage{algorithm}
\usepackage{algorithmic}
\usepackage{newfloat}
\usepackage{listings}
\floatstyle{ruled}
\newfloat{listing}{tb}{lst}{}
\floatname{listing}{Listing}

\setlist[itemize]{leftmargin=*,itemsep=-5pt}
\setlist[enumerate]{leftmargin=*,itemsep=-5pt}

\newcommand{\taskfullname}{Textual Reasoning about Actions and Change}
\newcommand{\taskname}{TRAC}

\title{\textsc{TRAC}: A Textual Benchmark for Reasoning about Actions and Change}

\author{
    Weinan He,
    Canming Huang, 
    Zhanhao Xiao, 
    Yongmei Liu \\
    Dept. of Computer Science, Sun Yat-sen University, Guangzhou 510006, China \\
    \texttt{\{heweinan, huangcm\}@mail2.sysu.edu.cn} \\
    \texttt{\{xiaozh9, ymliu\}@mail.sysu.edu.cn}
}

\begin{document}
\maketitle
\begin{abstract}

Reasoning about actions and change (RAC) is essential to understand and interact with the ever-changing environment. Previous AI research has shown the importance of fundamental and indispensable knowledge of actions, i.e., preconditions and effects. However, traditional methods rely on logical formalization which hinders practical applications. With recent transformer-based language models (LMs), reasoning over text is desirable and seemingly feasible, leading to the question of whether LMs can effectively and efficiently learn to solve RAC problems. We propose four essential RAC tasks as a comprehensive textual benchmark and generate problems in a way that minimizes the influence of other linguistic requirements (e.g., grounding) to focus on RAC. The resulting benchmark, \textbf{TRAC}, encompassing problems of various complexities, facilitates a more granular evaluation of LMs, precisely targeting the structural generalization ability much needed for RAC. Experiments with three high-performing transformers indicates that additional efforts are needed to tackle challenges raised by TRAC.

\end{abstract}


\section{Introduction}

Reasoning about actions and change (RAC) has been a central issue since the early days in AI \cite{McCarthy63SitCalc} and received much attention from the NLP community \cite{rw1Li20,rw4Zellers20,rw6Dalvi18}. Classical AI has long recognized the importance of \textbf{actions} and characterized the fundamental knowledge about actions: \textbf{preconditions} and \textbf{effects} \cite{Reiter01KIA}. Preconditions are the conditions that must be satisfied when actions are executed, while effects define the result. Consider the example where an agent is moving containers in a dock: Suppose a red container is on top of a green one, and the agent is given an instruction to \enquote{move the green container to another platform}. To achieve such a goal, it needs to know the requirements of moving a container (there is not any other container on top of it) and the effect of such actions.

While traditional approaches provide sound and effective reasoning by capturing preconditions and effects, they rely on expensive and difficult formalization. Consequently, inducing knowledge from text and reasoning directly is becoming more preferable. As transformers have shown their promising potential in many linguistic tasks \cite{RoBERTa,Raffel20T5}, we set out to investigate \textit{whether transformers can learn efficiently and effectively solve RAC problems \textit{over text}}.

Previous work involving actions in NLP is more \textit{application}-centric and usually contains one specific task (e.g. instruction following, prediction) \cite{rw1Li20,rw3Dan21,rw4Zellers20} without considering both preconditions and effects. Typically, additional linguistic requirements are highlighted instead of focusing on RAC.

We propose \textbf{\taskfullname} (\textbf{\taskname}), a comprehensive suite of four fundamental and granular RAC reasoning tasks, inspired by traditional reasoning problems \cite{KR04}. Recognizing the utmost importance of preconditions and effects, we include the two fundamental tasks, \textbf{Projection} and \textbf{Executability}, which \textit{directly target} the \textit{essential knowledge} of RAC. We also provide two composite tasks, \textbf{Planning} and \textbf{Goal-Recognition} for a more \textit{comprehensive} problem settings.
Together, the four aspects of \taskname{} enables a more granular evaluation. 
We position \taskname{} as a diagnostic benchmark that is a more direct and precise embodiment of RAC abilities: 

\begin{itemize}
	\item \textbf{Projection}: anticipate the effects of actions;
	\item \textbf{Executability}: decide if actions are applicable;
	\item \textbf{Planning}: decide if actions form a valid plan;
	\item \textbf{Goal-recognition}:recognize the goal from observations of actions.
\end{itemize}

We employ two principles in dataset construction. Firstly, we minimize the impact of other important abilities (e.g. grounding and language variance) for a \enquote{clean-room} evaluation that solely focuses on RAC. Secondly, we desire reasoning problems with controlled complexities for testing structural generalization. We design a framework that takes the action domain knowledge and the textual template as input, generates symbolic problems, and synthesizes the textual problems. In this paper, we select the blocks world \cite{BW}, a typical action domain for a proof-of-concept. However, the framework is domain-agnostic and is thus extensible to other action domains.

We design generalization tests based on the observation that once the classical AI systems are endowed with knowledge of actions, they can effectively solve structurally more complex problems. Therefore, we lay out four aspects of generalization experiments: 1) GE1 with more objects; 2) GE2 with longer action sequences; 3) GE3 with unseen names of objects; 4) GE4 with unseen conjunctive conditions. We train neural language models (LMs) of three high-performing transformer architectures, testing their efficiency and effectiveness on the \taskname{} datasets. The result shows that while transformers are able to induce and utilize knowledge from a large number of training examples, it remains a great challenge to efficiently generalize to structurally more complex problems on \taskname{}.


\section{Related Work}
Recent NLP tasks are embracing reasoning. While machine reading tasks such as SQuAD \cite{Squad} require inferencing ability to some extent, our work is the first one as we know to systematically study Textual RAC. Dynamic reasoning is also needed in commonsense tasks such as the Winograd Schema Challenge\cite{WSC,Superglue}. But environments in such tasks did not model well-defined actions and change. Compared to these tasks, our concern is formal reasoning that supports sound conclusions.

Our work is distinct from previous attempts of formal reasoning in natural language. Natural Logic studies valid inference over natural language \cite{NatLog}, but its recent developments focus on reasoning about semantic relations between lexical terms \cite{Angeli16}. Meanwhile, transformers have achieved promising results on textual deduction tasks \cite{Clark20,Saha20}. While they target general deduction such as \enquote{round people are nice; Bob is round; Thus Bob is nice}, we dedicate our work to RAC and requires systems to \textit{learn} the knowledge from examples.

Previous work has been exploring NLP tasks involving actions, especially with instruction following, action outcome prediction and procedural text generation tasks. \taskname{} is unique in its focus on fundamentals, broad coverage of reasoning tasks, and the granularity. Linguistic requirements (e.g. the ability of grounding and handling languge variance), though important, are foreign to RAC and are thus avoided. Compared to application-centric tasks, our work provide a more abstract view of tasks that embody the fundamental requirements of RAC. Moreover, this paper propose a suite of generalization tests that target structural complexities in RAC, which enables more fine-grained tests.

\noindent\textbf{Instruction Following (IF)}: Agents are given textual directives to execute actions in an environment. In \cite{rw7Zhou21,rw3Dan21,rw7Zhou21}, agents receive visual features or spatial coordinates as input. More importantly, IF does not target directly preconditions and effects as we do.

\noindent\textbf{Prediction}. Projection and prediction both concern the effect of actions. In \cite{rw4Zellers20}, symbolic attributes represent the world instead of language. Questions in bAbI \cite{Weston15} and ProPara \cite{rw6Dalvi18} ask about change of objects. However, they neglect the equally important preconditions.

\noindent\textbf{Procedure Generation}: Predicting the next instruction \cite{rw1Li20} or next step \cite{rw5bosselut2018simulating} might relate to the executability of actions, but they only cover partial RAC illustrated in this paper. While understanding the preconditions is needed, it could also implicate other extraneous factors such as the preferences of the instructor.

\noindent\textbf{Generalization} has always been a desirable trait. In \cite{rw3Dan21}, the authors  attacked IF systems with adversarial examples (e.g. with slight perturbations). Long-horizon problems are shown to be non-trivial for neural planners \cite{rw7Zhou21,rw8Xu19}. Our evaluation with transformers further corroborate the observation. Additionally, \taskname{} imposes more structural generalization requirements in a more granular way.

\section{\taskname}
\begin{table*}[tbh]
    \centering
    \resizebox{.9\linewidth}{!}{
    \begin{tabularx}{\linewidth}[c]{>{\centering\arraybackslash}l m{0.55\linewidth} X >{\centering\arraybackslash}c}
    \toprule
    \textbf{Task} &
    \multicolumn{1}{c}{\textbf{Context}} &
    \multicolumn{1}{c}{\textbf{Query}} &
    \multicolumn{1}{c}{\textbf{Answer}} \\
    \midrule
        
        \textbf{Projection} & \begin{tabular}{@{}p{\linewidth}} $s$: The green block is on the table. The red block is clear. The blue block is clear. The green block is clear. The red block is on the table. The blue block is on the table. \\ $\vec{a}$: Jane moves the green block from the table to the red block. \end{tabular} & $q$: The blue block is on top of the red block. & False \\
        
        \hline
        
        \textbf{EX} & \begin{tabular}{@{}p{\linewidth}} $s$: The olive block is on the table. The yellow block is on top of the olive block. The indigo block is clear. The indigo block is on top of the yellow block. \end{tabular} & $\vec{a}$: Jane moves the indigo block from the yellow block onto the table. & True \\
        
        \hline
        
        \textbf{Planning} & \begin{tabular}{@{}p{\linewidth}} $s$: The blue block is clear. The blue block is on top of the magenta block. The magenta block is on top of the white block. The white block is on the table.\\ $g$: the blue block is not on top of the magenta block \end{tabular} & $\vec{a}$:  Jane moves the blue block from the magenta block onto the table. & True \\
        
        \hline
        
        \textbf{GR} & \begin{tabular}{@{}p{\linewidth}} $s$: The blue block is clear. The blue block is on top of the magenta block. The magenta block is on top of the white block. The white block is on the table. \\ $\vec{a}$: Jane moves the blue block from the magenta block onto the table. \end{tabular} & $g$: the blue block is on top of the magenta block. & False \\
      
    \bottomrule
    \end{tabularx}
    }
    \caption{Examples of \taskname{} tasks. Each problem in \taskname{} consists of a context, a query, and an answer. The markers $s, \vec{a}, g$ are only shown here for reference. (EX=Executability, GR=Goal-Recognition)}
    \label{table:trac_task_example}
\end{table*}

In this section, we first introduce the theoretical background before we lay out the four tasks. Finally, we discuss dataset generation.

\subsection{Theoretical Preparation}
In RAC, the essential issues include understanding 1) \textit{the state of the environment}, 2) \textit{which actions are applicable}, and 3) \textit{how actions affect the environment}. For precise definitions of states and actions, we use the semantics of STRIPS \cite{STRIPS}, a typical formal language that enables sound reasoning. In STRIPS, an \textit{action domain} specifies the types of objects, predicates for describing the states, and how actions change the environment. In our exploration, we limit the action domains to be \textit{deterministic} and \textit{noise-free}. We also assume the \textit{unique name axioms}, where different objects have different names. 
\taskname{} is built upon the following concepts:

\begin{itemize}
\item A \textbf{state} is a set of atomic expressions (e.g., $\{ clear(A), on(A, B), \ldots \}$) that represents a snapshot of the environment at a specific time point. Atoms not in the state are considered false.
\item An \textbf{action} consists of four parts: name, precondition, add list and delete list, the last three of which are sets of atoms. An action is applicable to a state iff the atoms of the precondition are all contained in the state. When it is applied to a state $s$, expressions in the delete list will be removed from $s$ and those in the add list will be added into $s$, resulting in a new state $s'$.
\item An \textbf{action sequence} is an ordered sequence of actions. It is applicable to a state if and only if every action can be applied consecutively.
\item A \textbf{goal} is a ground formula that describes the objective state. We limit goals to be either a literal (an atom or its negation) or a conjunction of two. We only consider achievable goals.
\item A \textbf{plan} (wrt a goal and a state) is an applicable action sequence. It is \textit{goal-achieving} if the execution of the plan achieves the goal. A goal-achieve plan is \textit{goal-achieving} if no shorter plan is.
\end{itemize}

We use a variant of the blocks world (BW) as an example where 1) a table has infinite space to hold blocks, 2) blocks have identical sizes and can be stacked as towers (a block is either on another one or immediately on the table), and 3) a block can be moved only if certain conditions are met. Example \ref{example_bw} shows the predicates and actions in BW. Given an action domain and an initial state, the action and expression space can be determined, based on which the \taskname{} problems are generated.

\begin{exmp}
\textbf{Predicates in BW}:
\begin{enumerate}[topsep=-2pt]
              \item $on(x,y)$ states that $x$ is on $y$;
              \item $onTable(x)$ says $x$ is on the table;
              \item $clear(x)$ declares that there is no block on $x$.
\end{enumerate}
\label{example_bw}
\end{exmp}

\paragraph{Actions in BW.} We provide the definition of the action \(move(x, y, z)\) in the following. For the other definitions, refer to the appendix \ref{appx:actions_in_bw}.

\begin{exmp}
\textbf{Action $move(x, y, z)$}:
Move block $x$ that is on block $y$ onto block $z$.
\begin{itemize}[topsep=-2pt]
              \item Precondition: $clear(x), clear(z), on(x, y)$.
              \item Add list: $clear(y), on(x, z)$.
              \item Delete list: $clear(z), on(x, y)$.
\end{itemize}
\end{exmp}

\subsection{Reasoning Tasks in \taskname{}}
For a comprehensive and granular evaluation, we propose four different reasoning tasks, each focusing on an aspect of RAC. All four tasks in \taskname{} are formulated as text classification problems, similar to the deduction task \cite{Clark20} and Natural Language Inference task. Given the input of two texts, a \textit{context} and a \textit{query}, the system is asked to classify if the query is true accordingly. Concrete examples can be seen in Table \ref{table:trac_task_example}.

\paragraph{Projection.} The projection task directly asks about the effects of actions: Given an initial state $s$ and an applicable sequence $\vec{a}$ of $N$ actions, decide whether the projection query $q$, a proposition, would hold after the execution of $\vec{a}$. The context is $s$ and $\vec{a}$, and the query is $q$.

\paragraph{Executability.} This task directly targets the preconditions of actions: Given an initial state $s$ and a sequence $\vec{a}$ of $N$ actions, decide whether $\vec{a}$ can be executed consecutively in $s$. The context is $s$ and the query is $\vec{a}$.

\paragraph{Planning.} Planning is the task of formulating actions to fulfill a certain goal. In \taskname{}, we use the verification version that asks systems to recognize if the provided actions can achieve the goal: Given an initial state $s$ and a goal $g$, a proposition, and a sequence $\vec{a}$ of $N$ actions, decide if $a$ can achieve $g$. The context is $s$ and $g$, and the query is $\vec{a}$.

\paragraph{Goal-Recognition (GR).} GR is the task to recognize the goal from the partial observation of actions. We use a simplified version, where systems observe a partial action sequence and need to figure out if the given goal is the true objective: Given an initial state $s$, a potential goal $g$, and a sequence $\vec{a}$ of $N$ actions as the observation, decide if $g$ is the true objective. That is, decide if $\vec{a}$ is a prefix of any optimal plans to achieve $g$. The context is $s$ and $\vec{a}$, and the query is $g$.

\subsection{Dataset Generation}
We provide a framework to generate \taskname{} problems. It takes both the action domain and the language template as input to construct symbolic forms and translation, respectively.
For each task in \taskname{}, we generate three basic datasets, each having action sequences of different lengths (denoted as L1, L2, and L3 for lengths 1, 2, and 3 respectively). All examples in the datasets have $M=5$ objects. For the generalization tests discussed in detail in the next section, we also construct additional datasets with different parameters. Each problem is first generated in the symbolic form before being translated into textual form in English.

\paragraph{Symbolic Instance Generation.}
\noindent Commonly existing in all examples are the \textit{initial state} and the \textit{action sequence}, which serve as the foundation for all four tasks. While the initial state is always part of the context, the action sequence is either part of the context or is the query, depending on the task. Firstly we generate the \textit{state space} with $M$ objects according to the action domain. In the blocks world where blocks have different colors, their names (e.g. \enquote{the red block}) are randomly chosen from a pre-specified range. Secondly, the action space is computed, which includes all grounded possible actions w.r.t. each possible state. With these spaces, we construct the context and query for each \taskname{} task.

\paragraph{Projection.} The context includes an initial state and an action sequence. From the action space, we randomly sample $N$ actions to form a sequence that is executable in the initial state. For the query, we randomly generate a formula of the following form:
              \begin{equation}
                             l_1 \text{ or } l_1 \land l_2,
                             \label{eq:formula}
              \end{equation}
              where \(l_1\) and \(l_2\) are literals (atoms or their negations), e.g., \(onTable(Blue)\), \(\neg on(Green, Blue)\). The query is true if and only if it holds after executing the action sequence.
              
\paragraph{Executability.} The context is an initial state. For the query, we sample $N$ actions from the action space as a sequence. The query is true if and only if it is executable \textit{consecutively} in the initial state.

\paragraph{Planning.} The context consists of an initial state and a goal that is achievable with $N$ actions or less. The planning query is an action sequence of length $N$. Both the goal and the action sequence are generated at random. Goals share the same form as projection queries, shown in Formula \ref{eq:formula}. We only achievable goals in this task. The query is true if and only if it is a valid goal-achieving plan.

\paragraph{Goal-Recognition.} The context comprises an initial state and a sequence of $N$ actions as the partial observation. The GR query is a plausible goal of the same form as in the planning task. Both the action sequence and the goal are randomly generated. The query is true if and only if the observation is a prefix of any optimal plans that achieve the query.

\paragraph{Textual Form Synthesis.} We use templates to generate the actual problems, following the guideline of \enquote{clean-room} evaluation, which strives to focus on RAC instead of other linguistic requirements such as grounding and language variance. To maintain readability, our framework utilizes handcrafted templates for the specific action domain. These templates specify the translation of each predicate, action, goal, and projection query. A symbolic state will be converted into several sentences, each of which is a direct translation of the atomic expression. Similarly, an action sequence is translated into a concatenation of the action sentences. Projection queries are synthesized in the same fashion, but goals are processed differently to accommodate conjunctions with \enquote{and} if necessary. For example, the textual form of the goal $onTable(Blue) \land \neg on(Green, Blue)$ is \enquote{the blue block is on the table and the green block is not on the blue block}. 

As a result, we generate twelve datasets (four tasks, each with three datasets of various lengths of action sequences). Each dataset contains 15,000 label-balanced examples. We split the 15k examples into 12k training examples (where 2k are used as a dev set) and 3k testing examples.

\section{Experiments}
\label{section:exp}

\begin{table}[bht]
    \centering
    \begin{tabularx}{\columnwidth}{c p{1.8cm}<{\centering} p{1.8cm}<{\centering} p{1.8cm}<{\centering}}
        \toprule
        \textbf{Task} & \textbf{RoBERTa} & \textbf{GPT-2} & \textbf{T5} \\
        \midrule
        \textbf{PR} & \makecell{ \textbf{87.36} \\ (0.0396) } & \makecell{ 85.13 \\ (0.0336) } & \makecell{ 82.99 \\ (0.0227)} \\ \hline 
        \textbf{EX} & \makecell{ 99.73 \\ (0.0013) } & \makecell{ 99.37 \\ (0.0037) } & \makecell{ \textbf{98.83} \\ (0.0024)} \\ \hline 
        \textbf{PL} & \makecell{ 87.63 \\ (0.0158) } & \makecell{ \textbf{90.09} \\ (0.0157) } & \makecell{ 87.73 \\ (0.0110)} \\ \hline 
        \textbf{GR} & \makecell{ 96.82 \\ (0.0044) } & \makecell{ \textbf{97.44} \\ (0.0021) } & \makecell{ 94.04 \\ (0.0082)} \\
        \bottomrule
    \end{tabularx}
    \caption{Accuracies (percent signs omitted) and standard deviations (in parentheses) of the baselines on \taskname{}. Each cell corresponds to the model trained and tested on the specific dataset (column header) for the task. (PR=Projection, EX=Executability, PL=Planning, GR=Goal-Recognition)}
    \label{table:ind-acc}
\end{table}

We conduct experiments to address the following questions:
\begin{enumerate}
	\item Can transformers induce knowledge to effectively solve \taskname{} problems?
	\item Can they generalize to problems that are structurally more complex?
	\item How data-efficient are they?
\end{enumerate}
The datasets, code, and hyper-parameters are available in the supplementary materials.

\begin{table*}[htb]\fontsize{9}{10}
    \centering
    \begin{tabular}{ l|c|c|c|c|c|c|c }
        \toprule
        \multirow{2}{*}{\textbf{Task}} &
        \multirow{2}{*}{\textbf{SD}} &
        \multirow{2}{*}{\textbf{GE1}} &
        \multicolumn{2}{c|}{\textbf{GE2}} &
        \multirow{2}{*}{\textbf{GE3}} &
        \multicolumn{2}{c}{\textbf{GE4}} \\
            \cline{4-5} \cline{7-8}
            & & & \textbf{L4} & \textbf{L5} & &  \textbf{Literals} & \textbf{Conj.}  \\
        \midrule
        \textbf{PR} &  \makecell{ 87.36 \\ (0.0396) } & \makecell{ 58.19 \\ (0.0185) } &  \makecell{ 71.91 \\ (0.0428) } & \makecell{ 69.82 \\ (0.0301) } &   \makecell{ 85.09 \\ (0.0308) } &   \makecell{ 93.15 \\ (0.0537) } & \makecell{ 72.89 \\ (0.0248) } \\ \hline
        \textbf{EX} &  \makecell{ 99.73 \\ (0.0013) } & \makecell{ 87.91 \\ (0.0469) } &  \makecell{ 82.54 \\ (0.0200) } & \makecell{ 79.76 \\ (0.0159) } &   \makecell{ 99.01 \\ (0.0014) } &   N/A                            & N/A                            \\ \hline
        \textbf{PL} &  \makecell{ 87.63 \\ (0.0158) } & \makecell{ 80.40 \\ (0.0461) } &  \makecell{ 61.67 \\ (0.0169) } & \makecell{ 56.27 \\ (0.0086) } &   \makecell{ 91.81 \\ (0.0181) } &   \makecell{ 98.11 \\ (0.0014) } & \makecell{ 68.63 \\ (0.0441) } \\ \hline
        \textbf{GR} &  \makecell{ 96.82 \\ (0.0044) } & \makecell{ 79.66 \\ (0.0475) } &  N/A                            & N/A                            &   \makecell{ 94.69 \\ (0.0040) } &   \makecell{ 99.99 \\ (0.0001) } & \makecell{ 73.91 \\ (0.0028) } \\ 
        \bottomrule
    \end{tabular}
    \caption{Accuracies and standard deviations of the RoBERTa-base models on generalization experiments. Results from Table \ref{table:ind-acc} are shown in the second column (SD=Standard) as a comparison for GE1, GE2, and GE3. The last two columns report results for GE4: the baselines are trained using examples with only literals. (PR=Projection, EX=Executability, PL=Planning, GR=Goal-Recognition)}
    \label{table:roberta-base-ge}
\end{table*}

\subsection{Baseline Models}
We use three different LMs as our baseline models, each with different architectures: RoBERTa \cite{RoBERTa}, GPT-2 \cite{radford19gpt2}, and T5 \cite{Raffel20T5}. These architectures compose transformer layers in various typical fashions:
\begin{enumerate}[topsep=-1pt]
    \item RoBERTa features transformer-encoder layers;
    \item GPT-2 contains transformer-decoder layers;
    \item T5 combines both encoder and decoder layers.
\end{enumerate}

Driving by the classification nature of \taskname{}, the first two baseline models are built by adding a linear layer upon the transformer layers for both RoBERTa and GPT-2. As for T5, we use its text-to-text pre-training objective to generate the labels.
The input for RoBERTa is organized as \texttt{<s> context </s> query </s>} where \texttt{<s>} and \texttt{</s>} are separator symbols, following previous work \cite{Clark20}.
We tarin these LMs to predict the truth of the query using the cross-entropy loss function. Accuracy is used as the metric for evaluation, and we report the mean values and standard deviations of the repeated experiments. Appendix~\ref{appx:tf} gives details of the LMs.

\subsection{Effectiveness of Transformers}
We first evaluate the effectiveness of the baseline models on \taskname{} problems of the same structural complexity. Both the training set and the test set contain problems with \(M=5\) objects and \(N\) actions (\(N \in \{1, 2, 3\}\), where the ratio of L1:L2:L3 is 1:1:1). For each architecture, we train a baseline model separately for each reasoning task, resulting in twelve such models, shown in \ref{table:ind-acc}. In this setting, transformers are given enough training data and are required to induce knowledge about actions and change from examples.

In Table \ref{table:ind-acc}, transformers have shown capable performances, with all accuracies above 80\%. While they excel at Executability and Goal-Recognition, there is considerable room for improvement on Projection and Planning. Although these different transformer architectures have their own wins on different tasks, they do have rather similar performances, demonstrating that the challenge of \taskname{} is universal for transformers.

\subsection{Structural Generalization Experiments}

Targeting structural generalization ability, we design four out-of-distribution tests. Novel test examples that are more structurally complex are generated to this end. For the first three generalization experiments (GEs), we re-use models trained on previous datasets (with complexities \(M=5, N \in \{1, 2, 3\}\)). Since the last GE requires additional training, both training and test sets are necessary. In total, we created twenty additional datasets beyond the normal ones. Table \ref{table:roberta-base-ge} shows the results for the RoBERTa-base models.

\paragraph{GE1: More Objects.} In this test, we examine whether the baselines can handle \taskname{} problems containing more objects  (blocks in the BW domain). We generate a new dataset for each reasoning task, all of which involve ten objects. We evaluate the baselines trained on the standard datasets.
The results show that baselines have significantly worse performances on GE1 datasets for all tasks, most notably in Projection. This is expected, as the GE1 datasets have longer state descriptions and more complicated states. Compared to the training examples, the state descriptions in GE1 datasets contain 6.1 more sentences with 52.1 more words on average. Nonetheless, the results indicate that the baselines do not generalize well to more objects.

\paragraph{GE2: More Actions.} Naturally, we are interested in the generalization to longer action sequences, as the length often plays a vital role in both formal reasoning tasks and natural language tasks \cite{rw7Zhou21, rw8Xu19}. For Projection, Executability, and Planning tasks, we generated the L4 and the L5 datasets that have action sequences of length four and five, respectively. Goal-Recognition is left out as there are not enough test examples in this setting. 
The results from Table \ref{table:roberta-base-ge} confirm that the length of the action sequence is a vital factor, as the accuracies degrade for longer sequences. It is more apparent for planning problems, with an appalling 40\% accuracy loss on L5. This observation further verifies the universality of the long-horizon problem \cite{rw8Xu19}, even for transformer-based LMs.

\paragraph{GE3: Unseen Names.} Changing the names of the objects is supposed to have negligible impact. The ability of generalization to unseen symbols is desirable in reasoning over both formal and natural languages. Seeing this, we substitute the names of objects for previously unseen ones, resulting in four more datasets. They differ from the standard ones only in the object names. 
The results from Table~\ref{table:roberta-base-ge} show minor differences between the performances as expected, demonstrating the capability of the LMs to generalize beyond unseen names.

\paragraph{GE4: Compositionality in Goals and Projection Queries.} Reasoning requires the capability of \textit{compositionality}: to understand or manipulate higher-level structures composed of known components. One such ability is to combine two conditions as a conjunction. If systems understand conditions $A$ and $B$ separately, they are expected to realize that the conjunction $A \land B$ is true iff \textit{both} are true. In \taskname{} tasks, the ideal testbeds are Projection, Planning, and Goal-recognition, where the projection queries and goals could be partitioned into literals and conjunctions. 

In this experiment, we train baselines using examples with only literals in conditions and see if they can handle the examples with conjunctions. Therefore, new training datasets are needed. For each of the three tasks, we generate GE4-literals, a dataset of 15k instances with only literals in target conditions (10k for training, 2k as dev set, and the other 3k for testing), along with GE4-conjunctions, a dataset of 3k problems with only conjunctions. After training the baselines on GE4-literals, we compare their performances on test sets of literals and conjunctions.

The results from Table \ref{table:roberta-base-ge} show that the baselines do not generalize well to conjunctions, losing more than 20\% accuracies on all tasks. Such a phenomenon suggests that compositionality is not trivial in \taskname{} for the transformers. It is also noteworthy that conjunctive conditions have various forms: While conjunctions are represented by two sentences in Projection, they are of the form \enquote{condition1 and condition2} in Planning and Goal-Recognition. This leads us to believe that the performance loss is not about the introduction of the surface form \enquote{and}, but about the conceptual understanding of conjunctive compositionality, without which the models cannot generalize well.

\subsection{Data Efficiency}
\begin{figure}
	\includegraphics[width=\columnwidth]{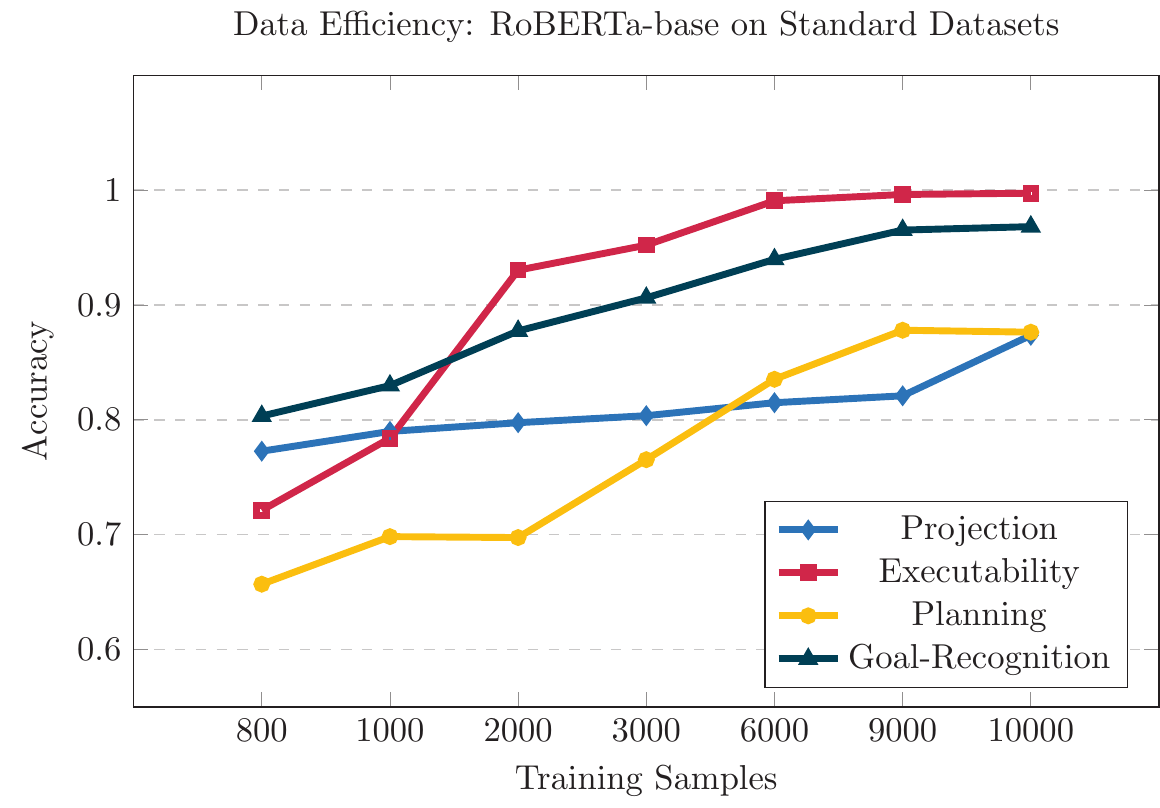}
	\caption{Accuracies of RoBERTa baselines vs sizes of training samples on the standard datasets of \taskname{}.}
	\label{fig:roberta_base_ratio_ind}
\end{figure}

\begin{figure}
	\includegraphics[width=\columnwidth]{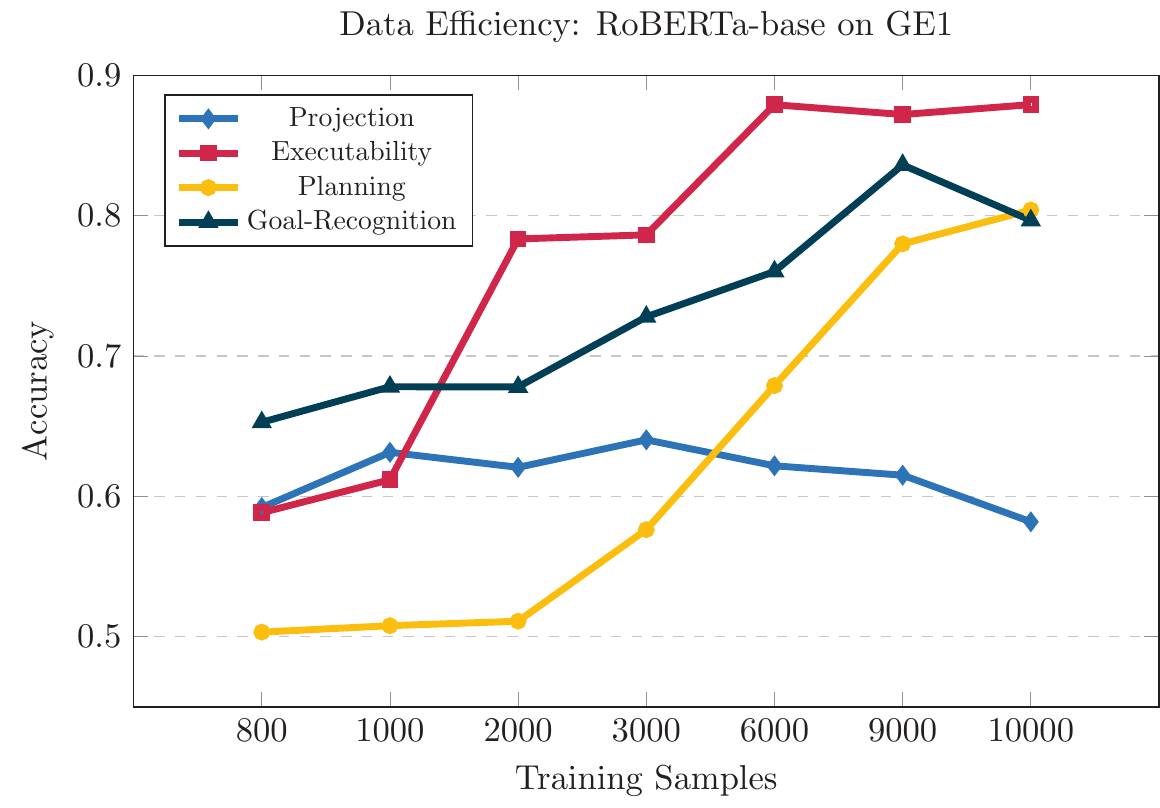}
	\caption{Accuracies of RoBERTa baselines vs sizes of training samples on the GE1 of \taskname{}.}
	\label{fig:roberta_base_ratio_ge1}
\end{figure}

\begin{figure}[h!]
	\includegraphics[width=\columnwidth]{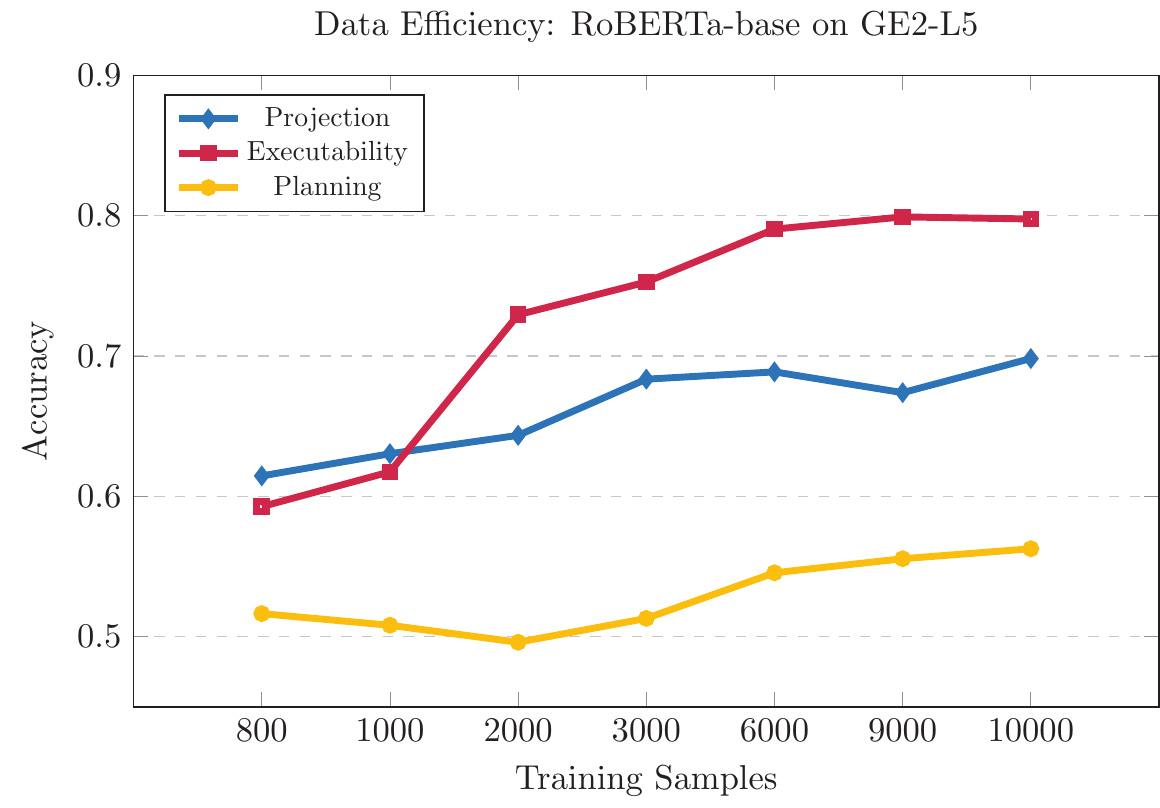}
	\caption{Accuracies of RoBERTa baselines vs sizes of training samples on the GE2 of \taskname{}.}
	\label{fig:roberta_base_ratio_ge2}
\end{figure}

In reality, humans need only a few examples to adapt to novel environments. To explore how many training samples are needed for the transformers, we plot the accuracies of the RoBERTa baselines with increasing numbers of training samples. In Figure~\ref{fig:roberta_base_ratio_ind}, we notice that the baselines require at least 3000 samples to have acceptable accuracies (above 80\%) on standard datasets. The inefficiency is even more obvious when it comes to GE1 and GE2 examples in Figure~\ref{fig:roberta_base_ratio_ge1} and Figure~\ref{fig:roberta_base_ratio_ge2}, respectively. Moreover, the planning task seems to be the most challenging one when training data is limited.

\paragraph{Other Transformers.}
We also evaluate GPT-2, T5, and a larger RoBERTa model on the GEs and the full results are provided in Appendix~\ref{appx:other_ges}.
\begin{itemize}
    \item GPT-2 and T5 have rather similar performances on GEs, suggesting that structural generalization is a universal challenge for transformers;
    \item Although the larger RoBERTa model outperforms smaller models, it also suffers when facing structurally more complex problems.
\end{itemize}

\section{Discussion}
\subsection{Transformers on \taskname{}}
Although we are optimistic about the future of transformers, their performances in the generalization tests indicate that transformers alone are not enough for RAC:
Firstly, the reasoning problems in this proof-of-concept evaluation involve few actions objects that would be quite effortless for humans. Secondly, the scale of generalization is rather minor (from three actions to four or five actions; from five objects to ten objects). Yet we could observe the struggle of transformers with such minute structurally complex problems.
As illustrated in \cite{rw1Li20}, transformers capture meaning in texts to some extent, which indicates that they have potential in modeling actions and change. Such potentials can also be seen in our experiments when transformers are given more than abundant training examples.
Meanwhile, \citeauthor{rw4Zellers20} showed that dedicated neural components other than transformers could help model state changes and predictions.
These lead us to the conjecture that additional mechanisms that model the \textbf{preconditions} and \textbf{effects} could be the next objectives towards solving \taskname{} problems.

\subsection{Broader Impact}

Humans rely on RAC to understand and interact with the dynamic reality. Our work serves as a first comprehensive characterization of the fundamental reasoning in text and contributes to
1) bypassing the need for complete formalization in dynamic reasoning tasks, towards a more practical usage in reality;
2) extensible framework to adopt other action domains without the need of complete redesigning the generation procedure;
3) providing a testbed for more generalization tests;
4) expanding the limits of transformers.

\subsection{Future Work}

\paragraph{More Complex Domains.} We choose the BW for its intuitive and simplistic nature (with one kind of object, three types of actions, and three predicates). Although the generalization experiments suffice currently to challenge transformers, real-world situations are more complicated. With the improvement of the algorithms, the need for a better arrangement of actions domains is emerging. In time, it could be beneficial to include several domains with various levels of complexities.

\paragraph{Balance Between Rigor and Natural.} Fo

\section{Conclusion}

In a time where language models excel at many natural language tasks, including deductive ones, we revisit the key reasoning abilities for dynamic worlds with actions and change. While preserving the essence of traditional formal reasoning, we set out to investigate how well transformers can reason rigorously over textual input, which avoids the need for a complete formalization of each specific problem. Using the semantics of STRIPS, we characterize four essential reasoning tasks about actions and change to form the \taskname{} benchmark. We devise a framework to generate symbolic problems and transform them into text, resulting in a suite of datasets of various complexities. We also design four further experiments that target different aspects of structural generalization. Built upon the high-performing transformers, the baselines are put to the test under different settings. Although they show promising results on in-distribution problems provided with more-than-abundant training examples, it is the out-of-distribution generalization tests that cause troubles. We argue that \taskname{} tasks could be used to 1) expand our understanding of the limitations of transformers and, 2) serve as a challenge for generalization in RAC over text. In the future, we expect to see more interesting work based on \taskname{}, such as better solvers with mechanisms to learn both preconditions and effects, and novel generalization tests that call for more specific reasoning abilities.


\bibliography{anthology,trac}

\begin{thebibliography}{24}
\expandafter\ifx\csname natexlab\endcsname\relax\def\natexlab#1{#1}\fi

\bibitem[{Angeli et~al.(2016)Angeli, Nayak, and Manning}]{Angeli16}
Gabor Angeli, Neha Nayak, and Christopher~D. Manning. 2016.
\newblock \href {https://doi.org/10.18653/v1/p16-1042} {Combining natural logic
  and shallow reasoning for question answering}.
\newblock In \emph{Proceedings of the 54th Annual Meeting of the Association
  for Computational Linguistics, {ACL} 2016, August 7-12, 2016, Berlin,
  Germany, Volume 1: Long Papers}. The Association for Computer Linguistics.

\bibitem[{Bosselut et~al.(2018)Bosselut, Ennis, Levy, Holtzman, Fox, and
  Choi}]{rw5bosselut2018simulating}
Antoine Bosselut, Corin Ennis, Omer Levy, Ari Holtzman, Dieter Fox, and Yejin
  Choi. 2018.
\newblock \href {https://openreview.net/forum?id=rJYFzMZC-} {Simulating action
  dynamics with neural process networks}.
\newblock In \emph{International Conference on Learning Representations}.

\bibitem[{Brachman and Levesque(2004)}]{KR04}
Ronald~J. Brachman and Hector~J. Levesque. 2004.
\newblock \href {https://doi.org/10.1016/B978-1-55860-932-7.X5083-3}
  {\emph{Knowledge Representation and Reasoning}}.
\newblock Elsevier.

\bibitem[{Clark et~al.(2020)Clark, Tafjord, and Richardson}]{Clark20}
Peter Clark, Oyvind Tafjord, and Kyle Richardson. 2020.
\newblock \href {https://doi.org/10.24963/ijcai.2020/537} {Transformers as soft
  reasoners over language}.
\newblock In \emph{Proceedings of the Twenty-Ninth International Joint
  Conference on Artificial Intelligence, {IJCAI-20}}, pages 3882--3890.
  International Joint Conferences on Artificial Intelligence Organization.
\newblock Main track.

\bibitem[{Cook and Liu(2002)}]{BW}
Stephen~A. Cook and Yongmei Liu. 2002.
\newblock \href {http://rutcor.rutgers.edu/\%7Eamai/aimath02/PAPERS/7.ps} {A
  complete axiomatization for blocks world}.
\newblock In \emph{International Symposium on Artificial Intelligence and
  Mathematics, AI{\&}M 2002, Fort Lauderdale, Florida, USA, January 2-4, 2002}.

\bibitem[{Dalvi et~al.(2018)Dalvi, Huang, Tandon, Yih, and Clark}]{rw6Dalvi18}
Bhavana Dalvi, Lifu Huang, Niket Tandon, Wen{-}tau Yih, and Peter Clark. 2018.
\newblock \href {https://doi.org/10.18653/v1/n18-1144} {Tracking state changes
  in procedural text: a challenge dataset and models for process paragraph
  comprehension}.
\newblock In \emph{Proceedings of the 2018 Conference of the North American
  Chapter of the Association for Computational Linguistics: Human Language
  Technologies, {NAACL-HLT} 2018, New Orleans, Louisiana, USA, June 1-6, 2018,
  Volume 1 (Long Papers)}, pages 1595--1604. Association for Computational
  Linguistics.

\bibitem[{Dan et~al.(2021)Dan, Zhou, and Roth}]{rw3Dan21}
Soham Dan, Michael Zhou, and Dan Roth. 2021.
\newblock \href {https://doi.org/10.18653/v1/2021.naacl-main.76}
  {Generalization in instruction following systems}.
\newblock In \emph{Proceedings of the 2021 Conference of the North American
  Chapter of the Association for Computational Linguistics: Human Language
  Technologies, {NAACL-HLT} 2021, Online, June 6-11, 2021}, pages 976--981.
  Association for Computational Linguistics.

\bibitem[{Fikes and Nilsson(1971)}]{STRIPS}
Richard Fikes and Nils~J. Nilsson. 1971.
\newblock \href {https://doi.org/10.1016/0004-3702(71)90010-5} {{STRIPS:} {A}
  new approach to the application of theorem proving to problem solving}.
\newblock \emph{Artif. Intell.}, 2(3/4):189--208.

\bibitem[{Lakoff(1970)}]{NatLog}
George Lakoff. 1970.
\newblock Linguistics and natural logic.
\newblock \emph{Synthese}, 22(1):151--271.

\bibitem[{Levesque et~al.(2012)Levesque, Davis, and Morgenstern}]{WSC}
Hector~J. Levesque, Ernest Davis, and Leora Morgenstern. 2012.
\newblock \href {http://www.aaai.org/ocs/index.php/KR/KR12/paper/view/4492}
  {The winograd schema challenge}.
\newblock In \emph{Principles of Knowledge Representation and Reasoning:
  Proceedings of the Thirteenth International Conference, {KR} 2012, Rome,
  Italy, June 10-14, 2012}. {AAAI} Press.

\bibitem[{Li et~al.(2021)Li, Nye, and Andreas}]{rw1Li20}
Belinda~Z. Li, Maxwell~I. Nye, and Jacob Andreas. 2021.
\newblock \href {https://doi.org/10.18653/v1/2021.acl-long.143} {Implicit
  representations of meaning in neural language models}.
\newblock In \emph{Proceedings of the 59th Annual Meeting of the Association
  for Computational Linguistics and the 11th International Joint Conference on
  Natural Language Processing, {ACL/IJCNLP} 2021, (Volume 1: Long Papers),
  Virtual Event, August 1-6, 2021}, pages 1813--1827. Association for
  Computational Linguistics.

\bibitem[{Liu et~al.(2019)Liu, Ott, Goyal, Du, Joshi, Chen, Levy, Lewis,
  Zettlemoyer, and Stoyanov}]{RoBERTa}
Yinhan Liu, Myle Ott, Naman Goyal, Jingfei Du, Mandar Joshi, Danqi Chen, Omer
  Levy, Mike Lewis, Luke Zettlemoyer, and Veselin Stoyanov. 2019.
\newblock \href {http://arxiv.org/abs/1907.11692} {Roberta: {A} robustly
  optimized {BERT} pretraining approach}.
\newblock \emph{CoRR}, abs/1907.11692.

\bibitem[{McCarthy(1963)}]{McCarthy63SitCalc}
John McCarthy. 1963.
\newblock Situations, actions, and causal laws.
\newblock Reprinted in Minsky69Book, pages 410--418.

\bibitem[{Radford et~al.(2019)Radford, Wu, Child, Luan, Amodei, Sutskever
  et~al.}]{radford19gpt2}
A.~Radford, J.~Wu, R.~Child, D.~Luan, D.~Amodei, I.~Sutskever, et~al. 2019.
\newblock Language models are unsupervised multitask learners.
\newblock \emph{OpenAI blog}, 1(8):9.

\bibitem[{Raffel et~al.(2020)Raffel, Shazeer, Roberts, Lee, Narang, Matena,
  Zhou, Li, and Liu}]{Raffel20T5}
Colin Raffel, Noam Shazeer, Adam Roberts, Katherine Lee, Sharan Narang, Michael
  Matena, Yanqi Zhou, Wei Li, and Peter~J. Liu. 2020.
\newblock \href {http://jmlr.org/papers/v21/20-074.html} {Exploring the limits
  of transfer learning with a unified text-to-text transformer}.
\newblock \emph{J. Mach. Learn. Res.}, 21:140:1--140:67.

\bibitem[{Rajpurkar et~al.(2018)Rajpurkar, Jia, and Liang}]{Squad}
Pranav Rajpurkar, Robin Jia, and Percy Liang. 2018.
\newblock \href {https://doi.org/10.18653/v1/P18-2124} {Know what you don't
  know: Unanswerable questions for squad}.
\newblock In \emph{Proceedings of the 56th Annual Meeting of the Association
  for Computational Linguistics, {ACL} 2018, Melbourne, Australia, July 15-20,
  2018, Volume 2: Short Papers}, pages 784--789. Association for Computational
  Linguistics.

\bibitem[{Reiter(2001)}]{Reiter01KIA}
Raymond Reiter. 2001.
\newblock \href {https://doi.org/10.7551/mitpress/4074.001.0001}
  {\emph{{K}nowledge in {A}ction: Logical Foundations for Specifying and
  Implementing Dynamical Systems}}.
\newblock The {MIT} Press.

\bibitem[{Saha et~al.(2020)Saha, Ghosh, Srivastava, and Bansal}]{Saha20}
Swarnadeep Saha, Sayan Ghosh, Shashank Srivastava, and Mohit Bansal. 2020.
\newblock \href {https://doi.org/10.18653/v1/2020.emnlp-main.9} {Prover: Proof
  generation for interpretable reasoning over rules}.
\newblock In \emph{Proceedings of the 2020 Conference on Empirical Methods in
  Natural Language Processing, {EMNLP} 2020, Online, November 16-20, 2020},
  pages 122--136. Association for Computational Linguistics.

\bibitem[{Wang et~al.(2019)Wang, Pruksachatkun, Nangia, Singh, Michael, Hill,
  Levy, and Bowman}]{Superglue}
Alex Wang, Yada Pruksachatkun, Nikita Nangia, Amanpreet Singh, Julian Michael,
  Felix Hill, Omer Levy, and Samuel~R. Bowman. 2019.
\newblock \href
  {https://proceedings.neurips.cc/paper/2019/hash/4496bf24afe7fab6f046bf4923da8de6-Abstract.html}
  {Superglue: {A} stickier benchmark for general-purpose language understanding
  systems}.
\newblock In \emph{Advances in Neural Information Processing Systems 32: Annual
  Conference on Neural Information Processing Systems 2019, NeurIPS 2019,
  December 8-14, 2019, Vancouver, BC, Canada}, pages 3261--3275.

\bibitem[{Weston et~al.(2016)Weston, Bordes, Chopra, and Mikolov}]{Weston15}
Jason Weston, Antoine Bordes, Sumit Chopra, and Tom{\'{a}}s Mikolov. 2016.
\newblock \href {http://arxiv.org/abs/1502.05698} {Towards ai-complete question
  answering: {A} set of prerequisite toy tasks}.
\newblock In \emph{4th International Conference on Learning Representations,
  {ICLR} 2016, San Juan, Puerto Rico, May 2-4, 2016, Conference Track
  Proceedings}.

\bibitem[{Wolf et~al.(2020)Wolf, Debut, Sanh, Chaumond, Delangue, Moi, Cistac,
  Rault, Louf, Funtowicz, Davison, Shleifer, von Platen, Ma, Jernite, Plu, Xu,
  Scao, Gugger, Drame, Lhoest, and Rush}]{wolf20tf}
Thomas Wolf, Lysandre Debut, Victor Sanh, Julien Chaumond, Clement Delangue,
  Anthony Moi, Pierric Cistac, Tim Rault, Rémi Louf, Morgan Funtowicz, Joe
  Davison, Sam Shleifer, Patrick von Platen, Clara Ma, Yacine Jernite, Julien
  Plu, Canwen Xu, Teven~Le Scao, Sylvain Gugger, Mariama Drame, Quentin Lhoest,
  and Alexander~M. Rush. 2020.
\newblock \href {https://www.aclweb.org/anthology/2020.emnlp-demos.6}
  {Transformers: State-of-the-art natural language processing}.
\newblock In \emph{Proceedings of the 2020 Conference on Empirical Methods in
  Natural Language Processing: System Demonstrations}, pages 38--45, Online.
  Association for Computational Linguistics.

\bibitem[{Xu et~al.(2019)Xu, Mart{\'{\i}}n{-}Mart{\'{\i}}n, Huang, Zhu,
  Savarese, and Fei{-}Fei}]{rw8Xu19}
Danfei Xu, Roberto Mart{\'{\i}}n{-}Mart{\'{\i}}n, De{-}An Huang, Yuke Zhu,
  Silvio Savarese, and Li~Fei{-}Fei. 2019.
\newblock \href
  {https://proceedings.neurips.cc/paper/2019/hash/3a835d3215755c435ef4fe9965a3f2a0-Abstract.html}
  {Regression planning networks}.
\newblock In \emph{Advances in Neural Information Processing Systems 32: Annual
  Conference on Neural Information Processing Systems 2019, NeurIPS 2019,
  December 8-14, 2019, Vancouver, BC, Canada}, pages 1317--1327.

\bibitem[{Zellers et~al.(2021)Zellers, Holtzman, Peters, Mottaghi, Kembhavi,
  Farhadi, and Choi}]{rw4Zellers20}
Rowan Zellers, Ari Holtzman, Matthew~E. Peters, Roozbeh Mottaghi, Aniruddha
  Kembhavi, Ali Farhadi, and Yejin Choi. 2021.
\newblock \href {https://doi.org/10.18653/v1/2021.acl-long.159} {Piglet:
  Language grounding through neuro-symbolic interaction in a 3d world}.
\newblock In \emph{Proceedings of the 59th Annual Meeting of the Association
  for Computational Linguistics and the 11th International Joint Conference on
  Natural Language Processing, {ACL/IJCNLP} 2021, (Volume 1: Long Papers),
  Virtual Event, August 1-6, 2021}, pages 2040--2050. Association for
  Computational Linguistics.

\bibitem[{Zhou et~al.(2021)Zhou, Yin, and Neubig}]{rw7Zhou21}
Shuyan Zhou, Pengcheng Yin, and Graham Neubig. 2021.
\newblock \href {http://arxiv.org/abs/2109.08214} {Hierarchical control of
  situated agents through natural language}.
\newblock \emph{CoRR}, abs/2109.08214.

\end{thebibliography}

\clearpage
\appendix
\label{sec:appendix}

\section{Experiment Details}

\begin{table*}[thb]\fontsize{9}{10}
    \centering
	\begin{tabular}{ l|c|c|c|c|c|c|c }
		\toprule
		\multirow{2}{*}{\textbf{Task}} &
		\multirow{2}{*}{\textbf{SD}} &
		\multirow{2}{*}{\textbf{GE1}} &
		\multicolumn{2}{c|}{\textbf{GE2}} &
		\multirow{2}{*}{\textbf{GE3}} &
		\multicolumn{2}{c}{\textbf{GE4}} \\
		\cline{4-5} \cline{7-8} & & & \textbf{L4} & \textbf{L5} & &  \textbf{Literals} & \textbf{Conj.}  \\
		\midrule
        \textbf{PR} & \makecell{ 98.79 \\ (0.0172) } & \makecell{ 57.81 \\ (0.0238) } &  \makecell{ 94.25 \\ (0.0583) } &  \makecell{ 88.16 \\ (0.0830) } &  \makecell{ 97.60 \\ (0.0228) } & \makecell{ 100.00 \\ (0.0000) } &   \makecell{ 74.41 \\ (0.0363) } \\ \hline       
        \textbf{EX} & \makecell{ 99.85 \\ (0.0007) } & \makecell{ 96.19 \\ (0.0128) } &  \makecell{ 89.97 \\ (0.0132) } &  \makecell{ 86.33 \\ (0.0187) } &  \makecell{ 99.48 \\ (0.0011) } & N/A                             &   N/A                            \\ \hline    
        \textbf{PL} & \makecell{ 93.94 \\ (0.0099) } & \makecell{ 84.69 \\ (0.0846) } &  \makecell{ 63.35 \\ (0.0270) } &  \makecell{ 56.69 \\ (0.0215) } &  \makecell{ 96.61 \\ (0.0051) } & \makecell{ 98.65 \\ (0.0037) }  &   \makecell{ 72.84 \\ (0.0839) } \\ \hline       
        \textbf{GR} & \makecell{ 98.70 \\ (0.0014) } & \makecell{ 88.65 \\ (0.0170) } &  N/A                            &  N/A                            &  \makecell{ 97.65 \\ (0.0044) } & \makecell{ 100.00 \\ (0.0000) } &   \makecell{ 74.07 \\ (0.0050) } \\      
		\bottomrule
	\end{tabular}
    \caption{Accuracies and standard deviations of the RoBERTa-large models on generalization experiments. (SD=Standard Dataset, PR=Projection, EX=Executability, PL=Planning, GR=Goal-Recognition)}
    \label{table:roberta-large-ge}
\end{table*}

\begin{table*}[thb]\fontsize{9}{10}
    \centering
	\begin{tabular}{ l|c|c|c|c|c|c|c }
		\toprule
		\multirow{2}{*}{\textbf{Task}} &
		\multirow{2}{*}{\textbf{SD}} &
		\multirow{2}{*}{\textbf{GE1}} &
		\multicolumn{2}{c|}{\textbf{GE2}} &
		\multirow{2}{*}{\textbf{GE3}} &
		\multicolumn{2}{c}{\textbf{GE4}} \\
		\cline{4-5} \cline{7-8} & & & \textbf{L4} & \textbf{L5} & &  \textbf{Literals} & \textbf{Conj.}  \\
		\midrule
		\textbf{PR} & \makecell{ 85.13 \\ (0.0336)} & \makecell{ 68.73 \\ (0.0397) } & \makecell{ 70.75 \\ (0.0443) } & \makecell{ 69.73 \\ (0.0447) } & \makecell{ 83.79 \\ (0.0257) } & \makecell{ 89.08 \\ (0.0524) } & \makecell{ 66.49 \\ (0.0280)  } \\ \hline
        \textbf{EX} & \makecell{ 99.37 \\ (0.0037)} & \makecell{ 90.95 \\ (0.0350) } & \makecell{ 89.84 \\ (0.0331) } & \makecell{ 88.11 \\ (0.0303) } & \makecell{ 97.40 \\ (0.0103) } & N/A                            & N/A                             \\ \hline
        \textbf{PL} & \makecell{ 90.09 \\ (0.0157)} & \makecell{ 79.45 \\ (0.0359) } & \makecell{ 61.93 \\ (0.0229) } & \makecell{ 57.13 \\ (0.0172) } & \makecell{ 91.97 \\ (0.0171) } & \makecell{ 96.75 \\ (0.0238) } & \makecell{ 62.83 \\ (0.0194)  } \\ \hline
        \textbf{GR} & \makecell{ 97.44 \\ (0.0021)} & \makecell{ 91.12 \\ (0.0121) } & N/A                            & N/A                            & \makecell{ 94.84 \\ (0.0074) } & \makecell{ 100.00 \\ (0.0000)} & \makecell{ 73.47 \\ (0.0099)  } \\ 
		\bottomrule
	\end{tabular}
    \caption{Accuracies and standard deviations of the GPT-2 models on generalization experiments. (SD=Standard Dataset, PR=Projection, EX=Executability, PL=Planning, GR=Goal-Recognition)}
    \label{table:gpt2-ge}
\end{table*}

\begin{table*}[hbt]\fontsize{9}{10}
    \centering
	\begin{tabular}{ l|c|c|c|c|c|c|c }
		\toprule
		\multirow{2}{*}{\textbf{Task}} &
		\multirow{2}{*}{\textbf{SD}} &
		\multirow{2}{*}{\textbf{GE1}} &
		\multicolumn{2}{c|}{\textbf{GE2}} &
		\multirow{2}{*}{\textbf{GE3}} &
		\multicolumn{2}{c}{\textbf{GE4}} \\
		\cline{4-5} \cline{7-8} & & & \textbf{L4} & \textbf{L5} & &  \textbf{Literals} & \textbf{Conj.}  \\
		\midrule
    \textbf{PR} & \makecell{ 82.99 \\ (0.0227) } & \makecell{ 68.35 \\ (0.0132) } & \makecell{ 67.69 \\ (0.0188) } & \makecell{ 66.80 \\ (0.0163) } & \makecell{ 81.34 \\ (0.0230) } & \makecell{ 83.27 \\ (0.0237) } & \makecell{ 71.35 \\ (0.0082) } \\ \hline
    \textbf{EX} & \makecell{ 98.83 \\ (0.0024) } & \makecell{ 89.36 \\ (0.0308) } & \makecell{ 81.77 \\ (0.0222) } & \makecell{ 80.15 \\ (0.0155) } & \makecell{ 98.12 \\ (0.0014) } & N/A                            & N/A                            \\ \hline
    \textbf{PL} & \makecell{ 87.73 \\ (0.0110) } & \makecell{ 81.03 \\ (0.0588) } & \makecell{ 68.25 \\ (0.0127) } & \makecell{ 61.92 \\ (0.0142) } & \makecell{ 89.49 \\ (0.0137) } & \makecell{ 97.72 \\ (0.0038) } & \makecell{ 65.68 \\ (0.0286) } \\ \hline
    \textbf{GR} & \makecell{ 94.04 \\ (0.0082) } & \makecell{ 82.74 \\ (0.0366) } & N/A                            & N/A                            & \makecell{ 90.61 \\ (0.0134) } & \makecell{ 99.95 \\ (0.0005) } & \makecell{ 81.32 \\ (0.0075) } \\
		\bottomrule
	\end{tabular}
    \caption{Accuracies and standard deviations of the T5 models on generalization experiments. (SD=Standard Dataset, PR=Projection, EX=Executability, PL=Planning, GR=Goal-Recognition)}
    \label{table:t5-ge}
\end{table*}

\subsection{GEs for RoBERTa-large, GPT-2, and T5}
\label{appx:other_ges}
The results for GEs for RoBERTa-large, GPT-2, and T5 can be seen from Table \ref{table:roberta-large-ge}, Table \ref{table:gpt2-ge}, and Table \ref{table:t5-ge} respectively, which suggests:
\begin{itemize}[topsep=-2pt]
    \item Number of parameters matters. Compare the results of RoBERTa-base and RoBERTa-large, we can clearly see that the larger models outperform the smaller ones consistently. However, training larger LMs is notoriously more time-consuming and expensive. Additionally, larger models hinder practical applications in the real world where inference time is typically short.
    \item The overall generalization performances are similar for different transformer architectures. As the parameters of models (RoBERTa-base, GPT-2 and T5) are at the same level, their struggle facing the generalization tests are alike.
\end{itemize}

The data efficiency test for GE2 shows that transformers plateau once given more-than-abundant training examples.

\subsection{Computing Infrastructure}
We use a workstation with Intel i9-10980XE CPU (16 cores), 128GiB of RAM, and  RTX 3090 GPU. The experiments took about 100 hours.

\subsection{License of \taskname{}}
The datasets and the code are released under the CRAPL license (the Community Research and Academic Programming License). The full text of the license is included in the supplmentary material.

\subsection{Transformers}
\label{appx:tf}
We use the HuggingFace \texttt{transformers} \cite{wolf20tf} implementation.
The following shows the LMs with the number of parameters:
\begin{itemize}
    \item RoBERTa-base: 125M;
    \item GPT-2-small: 117M;
    \item T5-base: 220M;
    \item RoBERTa-large: 770M.
\end{itemize}

\noindent The following hyper-parameters are used:
\begin{itemize}[topsep=-2pt]
    \item The learning rate is 1e-5 for RoBERTa-base and 1e-4 for the others;
    \item The maximum sequence length is 256;
    \item The batch size is 16;
    \item The weight decay is 0.01;
    \item The warmup ratio is 0.06.
\end{itemize}
We repeated every experiment five times to calculate the mean values and standard deviations using five different seeds.

\noindent We list the input-output examples in Listing~\ref{lst:tfio}.

\section{The Blocks World}
\subsection{Actions in The Blocks World}
\label{appx:actions_in_bw}
We provide the other two actions of BW.
\noindent $moveToTable(x, y)$: Move block $x$ that is on block $y$ onto the table.
\begin{itemize}
	\item Precondition: $clear(x), on(x, y).$
	\item Add list: $onTable(x), clear(y)$.
	\item Delete list: $on(x, y)$.
\end{itemize}

\noindent $moveFromTable(x, y)$: Move block $x$ that is on the table onto block $y$.
\begin{itemize}
	\item Precondition: $onTable(x), clear(x), clear(y)$.
	\item Add list: $on(x, y)$.
	\item Delete list: $onTable(x), clear(y)$.
\end{itemize}

\subsection{Symbolic Forms}
We list the symbolic forms of both the BW domain and the \taskname{} examples in the paper.

\begin{itemize}[topsep=-2pt]
\item Symbolic forms of the problems in Table \ref{table:trac_task_example} are shown in Listing~\ref{lst:taskexamples}.
\item The action domain BW is defined in a PDDL file. Its content is shown in Listing~\ref{lst:bw}.
\end{itemize}

\begin{listing}[htb]%
\caption{\taskname{} Symbolic Examples}%
\label{lst:taskexamples}%
\begin{lstlisting}
Projection Example
Initial state = {
	onTable(Green),
	clear(Red),
	clear(Blue),
	clear(Green),
	onTable(Red),
	onTable(Blue)
}
Action sequence = [
    moveFromTable(Green, Red)
]
Query = on(Blue, Red)

Executability Example
Initial state = {
	onTable(Olive),
	on(Yellow, Olive),
	clear(Indigo),
	on(Indigo, Yellow)
}
Query = [ 
    moveToTable(Indigo, Yellow)
]

Planning Example
Initial state = {
	clear(Blue),
	on(Blue, Magenta),
	on(Magenta, White),
	onTable(White)
}
Goal = !on(Blue, Magenta)
Query = [ 
    moveToTable(Blue, Magenta)
]

Goal-Recognition Example
Initial state = {
	clear(Blue),
	on(Blue, Magenta),
	on(Magenta, White),
	onTable(White)
}
Action Sequence = [ 
    moveToTable(Blue, Magenta)
]
Query = on(Blue, Magenta)
\end{lstlisting}
\end{listing}

\begin{listing}[thb] 
\caption{The action domain BW in PDDL}%
\label{lst:bw}%
\begin{lstlisting}
(define 
	(domain blocksworld)
	(:requirements :strips :typing)
	(:types block - object)
	(:predicates 	(clear ?x - block) 
					(on ?x - block ?y - block)
					(ontable ?x - block))
	(:action move
		:parameters (?x - block ?y - block  ?z - block)
		:precondition (and (clear ?x)
						   (clear ?z)
						   (on ?x ?y))
		:effect (and (not (clear ?z)) 
					 (not (on ?x ?y))
					 (on ?x ?z)
					 (clear ?y)))
	
	(:action movetotable
		:parameters (?x - block ?y - block)
		:precondition (and (clear ?x)
						   (on ?x ?y))
		:effect (and (not (on ?x ?y))
					 (clear ?y)
					 (ontable ?x)))

	(:action movefromtable
		:parameters (?x - block ?y - block)
		:precondition (and (ontable ?x)
						   (clear ?x)
						   (clear ?y))
		:effect (and (not (ontable ?x))
					 (not (clear ?y))
					 (on ?x ?y)))
)
\end{lstlisting}
\end{listing}

\begin{listing}[thb] 
\caption{Input And Output for Transformers}%
\label{lst:tfio}%
\begin{lstlisting}
RoBERTa
Input: "<s> The yellow block is on the table. The magenta block is on top of the pink block. The gray block is clear. The gray block is on the table. The magenta block is clear. The pink block is on top of the green block. The green block is on the table. The yellow block is clear. Jane moves the yellow block from the table to the gray block. </s> The green block is clear. The gray block is not on top of the yellow block. </s>"
Output: 0

GPT-2
Input: "The yellow block is on the table. The magenta block is on top of the pink block. The gray block is clear. The gray block is on the table. The magenta block is clear. The pink block is on top of the green block. The green block is on the table. The yellow block is clear. Jane moves the yellow block from the table to the gray block. The green block is clear. The gray block is not on top of the yellow block."
Output: 0

T5
Input: (Same as GPT-2)
Output: "No"
\end{lstlisting}
\end{listing}

\end{document}